# Tradeoffs in Resampling and Filtering for Imbalanced Classification

**Ryan Muther and David Smith**
Northeastern University
Boston, MA
muther.r@northeastern.edu, dasmith@ccs.neu.edu

## Abstract

Imbalanced classification problems are extremely common in natural language processing and are solved using a variety of resampling and filtering techniques, which often involve making decisions on how to select training data or decide which test examples should be labeled by the model. We examine the tradeoffs in model performance involved in choices of training sample and filter training and test data in heavily imbalanced token classification task and examine the relationship between the magnitude of these tradeoffs and the base rate of the phenomenon of interest. In experiments on sequence tagging to detect rare phenomena in English and Arabic texts, we find that different methods of selecting training data bring tradeoffs in effectiveness and efficiency. We also see that in highly imbalanced cases, filtering test data using first-pass retrieval models is as important for model performance as selecting training data. The base rate of a rare positive class has a clear effect on the magnitude of the changes in performance caused by the selection of training or test data. As the base rate increases, the differences brought about by those choices decreases.

## 1 Introduction

Many linguistic phenomena and document genres are rare and so, in many classification and tagging tasks, a single unmarked or "background" label predominates. This imbalance is especially common in information retrieval and similar tasks, where a few relevant documents or passages can be arbitrarily outnumbered by millions of non-relevant documents.

In machine learning, this class-imbalance problem is often addressed by re-sampling or re-weighting the training set (Juba and Le, 2019). In practice, many tasks with extreme class imbalance also adopt different strategies at test time, as well— often ranking and evaluating sets of candidate outputs instead of classifying them individually. For example, in question answering, where answers to questions are rare, it was standard practice to use explicitly curated datasets like SQuAD consisting exclusively of positive examples of questions and the correct responses to those questions. More recent work like SQuAD 2.0 (Rajpurkar et al, 2018) has broadened the scope of such corpora to include unanswerable but plausible questions (i.e. near-miss negative examples) in training and test sets. In theory, one could generate a massive number of negative examples by sampling answers and pairing them with random spans of text, but most such examples are uninformative for the purposes of modeling. In contrast, for tasks such as named entity recognition, where the phenomenon of interest occurs in most documents, it is common to train and test models on the entirety of a collection. This brings with it the implicit assumption that the majority of the negative examples are helpful for training the model to differentiate named entities from background noise.

This naturally leads one to question how useful negative examples are in detecting linguistic phenomena. While they are obviously useful in cases where the phenomenon of interest is common, their usefulness for rarer phenomena is not precisely known. Additionally, the relationship between the rarity of the phenomenon in question and the effects of different methods for constructing a set of negative examples is also unclear. Besides model accuracy, we also consider efficiency. Training on a smaller numbers of negative examples clearly takes less time, and filtering out "obvious" negative samples using a simple first-pass retrieval model makes inference at test time more efficient.

In this paper, we examine these relationships by studying two example tasks in different languages with different base rates as case studies. In particular, we will attempt to solve the problems of detecting the start points of obituaries in 19th century American newspapers in English and those of citations, hereafter referred to by their Arabic name *isnads*, in classical Arabic texts from the 6th to 11th centuries, which unlike modern scientific citations tend to be integrated into running text and difficult to distinguish based on layout cues alone. We choose to use these tasks as they exhibit problems with significant class imbalance which we can use to study these relationships, yet are simpler to evaluate than problems like question answering where span level evaluation metrics must be used.

This paper is organized as follows. In section 2, we will present an overview of related work. Section 3 will describe the data used in our experiments. Section 4 will present an overview of model architectures used in our experiments. Section 5 will present the results of our experiments, and section 6 will discuss broader implications of these results and possible avenues of future work.

## 2 Related Work

This work is closely related to work on understanding imbalanced classification problems. Juba and Le (2019), for example, posit that classifier performance in terms of precision and recall in highly unbalanced settings is limited unless one has access to an often impractically high amount of data, which they demonstrate using experiments in statistical machine translation. However, in many domains, additional data may be difficult or impossible to acquire, so other solutions to the problem must be used, such as oversampling the positive class, undersampling the negative class (He and Garcia, 2009), or cost-sensitive classification (Ting, 2002).

As discussed above, different problems have different norms regarding how to deal with the problem of class imbalance, often related to the rarity of the phenomenon of interest, as typified by the difference between question answering and named entity recognition. It is also common to use a retrieval model as a first step in the process of creating datasets, such as the Microsoft Research Paraphrase Corpus (Dolan and Brockett, 2005). While these retrieval models are usually not modified after the (biased) sample of passages has been selected, Yang et al. (2019) find some benefits for question answering for improving a first-stage retrieval model given a fixed, second-stage answer-extraction model. This previous work focuses on these individual strategies, either at training or test time; as Juba and Le (2019) demonstrate, the effectiveness of strategies to mitigate class imbalance depend heavily on the base rate of the minority class. In this paper, we explore in a unified framework the tradeoffs in end-to-end performance among different sampling strategies for the training data, data filtering with retrieval at training and test time, and different training methods for the final classifier.

## 3 Data and Preprocessing

The data for these experiments comes two sources: the Richmond Daily Dispatch archives from 1860-65 (Retting, 2007) and the Open Islamicate Texts Initiative (OpenITI) corpus (Romanov and Seydi, 2019.)

The Richmond Daily Dispatch archives is comprised of 4,086 manually transcribed newspaper pages. For modeling purposes, we case fold and treat punctuation as separate tokens. Since many sequence-to-sequence models struggle when presented with long texts, we consider a single document to be a column of text as defined in the raw XML files of the transcripts, giving 28,040 columns which contain 28,678,804 tokens in total, and contain 4,934 obituaries.

The OpenITI collection consists of 4,285 transcribed texts in classical Arabic collected from digital libraries by scholars from the University of Vienna and University of Leipzig totaling 1.5 billion words. As with the Richmond corpus, the individual documents in the OpenITI corpus are complete texts and are often quite long, with the longest single text containing over 112 million words. The texts in the corpus are largely unpunctuated, and what punctuation there is is a modern editorial intervention, so we have removed punctuation from the texts when it exists. As such, it is difficult to break the texts down into smaller units than complete texts. We have also performed orthographic normalization to remove different variants of the same character so that the model isn't influenced by the orthographic choices of any particular author.

For these experiments, we have worked with a group of scholars in the field of premodern Islamic history to create a training dataset of tagged text from 36 labeled texts, with 149 distinct tagged regions of text. In total, the tagged text consists of 805,017 tokens containing 2,865 *isnads*.

# 4  Models

For our experiments, we treat the problem of locating the start points of obituaries and *isnads* as a two-stage tagging problem. This is done to avoid presenting complete documents to the sequence labeling models, which would perform poorly on the text of a whole column or tagged region of text, as both of these are sometimes as long as a thousand tokens. First, the documents are split into chunks of at most some fixed size (we use 100 and 75 for the Richmond and OpenITI datasets respectively, our reasoning will be explained below) to create initial training and test data for the sequence tagging models. These datasets may then be sampled in one of several ways, which will be described in section 4.1, to create the training data for the tagging models. We then employ one of several different sequence-to-sequence models, namely conditional random fields (Lafferty et al, 2001) and bidirectional LSTMs (Hochreiter and Schmidhuber, 1997; Graves, 2005), which will be detailed in section 4.2.

## 4.1  Sampling model

As noted above, we look at several different ways of selecting possible training documents. The first is to use only the positive examples in the training set (that is, those chunks that contain the start points of obituaries and *isnads*) to train the model. Secondly, one could randomly sample negative examples from among the rest of the training set until to create a new training set with the desired level of precision. For our experiments, we sample negative samples so they the precision of the resultant training set is the same as that for the retrieval models discussed below. Additionally, one could use a retrieval model trained to retrieve the phenomenon of interest as a filter to select which documents from the training set to use as training data and which from among the test set to actually use the model to predict locations in. This is done by splitting the training data into two sets, the using each half in turn to train a retrieval model which is used to retrieve documents from the other half of the training data. The union of the two retrieved sets of documents is then used as training data for the tagging model. This has the effect of creating a training set of positive and near-miss negative samples where the retrieval model mistook an uninteresting region for one of interest. This focuses the model on the more difficult decisions nearer to the model's decision boundary rather than giving the model uninteresting and uninformative examples that obviously contain no points of interest. In our experiments, we use a simple bag-of-ngrams logistic regression model implemented using sklearn. To improve recall, we tune the threshold of the classifier so that it has optimal F2 (recall-weighted F-measure), trying various chunk sizes from {25,50,75,100} orders of ngram from {1,2,3}, and the presence or absence of TFIDF feature weighting. We found that the retrieval models performed best for the *isnads* using TFIDF-weighted bigram features with a chunk size of 75 and a threshold of .14, while the obit retrieval models optimal operating point was TFIDF-weighted bigram features with a chunk size of 100 and a threshold of .03. Finally, one could train the tagging model using all of the available data.

## 4.2  Tagging model

As noted above, we implement and test two different tagging models; CRFs using local features and LSTMs using embeddings trained on their respective corpora.

The CRFs were implemented using sklearn-crfsuite using counts within a fixed-size window on either side of the word to be tagged as well as the token to be tagged itself as features for the model. Thus, if the word "the" occurred twice before the token being tagged, the feature "the_before" would be created with the value 2 for that token. We also investigated using words that don't occur in the window (unsupported features) as features with value 0, but this created a large number of features that made model training too memory-intensive. The intuition behind these features is that the points in text we seek to identify share common contexts (i.e. those used to convey information in an obituary in the Richmond dataset or to describe the transmission of information between individuals in the OpenITI dataset) that these features will enable the model to learn.

The LSTM was implemented using AllenNLP and trained using Adam with the default parameters for a maximum of 500 epochs, though most models converged in less than 200. We use single layer networks with a hidden layer size of 100. For our initial word embeddings, we used 200-dimensions vectors created by fitting GloVe embeddings (Pennington, 2014) on the full Richmond and OpenITI corpora preprocessed as described in Section 3.

| Training Set | Model | P | R | F |
|---|---|---|---|---|
| Positive | CRF | .082 | .090 | .084 |
| RandomNeg | CRF | .138 | .146 | .140 |
| Retrieval | CRF | .822 | .779 | .787 |
| All | CRF | .798 | .727 | .711 |
| Positive | LSTM | .077 | .085 | .079 |
| RandomNeg | LSTM | .161 | .165 | .160 |
| Retrieval | LSTM | .828 | .750 | .772 |
| All | LSTM | .843 | .729 | .776 |

| Training Set | Model | P | R | F |
|---|---|---|---|---|
| Positive | CRF | 0.790 | 0.800 | 0.795 |
| RandomNeg | CRF | 0.800 | 0.802 | 0.801 |
| Retrieval | CRF | 0.822 | 0.779 | 0.800 |
| All | CRF | 0.835 | 0.730 | 0.779 |
| Positive | LSTM | 0.802 | 0.802 | 0.802 |
| RandomNeg | LSTM | 0.815 | 0.791 | 0.803 |
| Retrieval | LSTM | 0.828 | 0.751 | 0.788 |
| All | LSTM | 0.852 | 0.733 | 0.788 |

Table 1: Obituary retrieval results tested on **all documents** (1a, left) and on **retrieved documents** (1b, right)

| Training Set | Model | P | R | F |
|---|---|---|---|---|
| Positive | CRF | .355 | .412 | .355 |
| RandomNeg | CRF | .399 | .413 | .380 |
| Retrieval | CRF | .535 | .356 | .403 |
| All | CRF | .489 | .351 | .385 |
| Positive | LSTM | .301 | .401 | .320 |
| RandomNeg | LSTM | .352 | .368 | .337 |
| Retrieval | LSTM | .463 | .328 | .366 |
| All | LSTM | .469 | .328 | .391 |

| Training Set | Model | P | R | F |
|---|---|---|---|---|
| Positive | CRF | 0.442 | 0.436 | 0.439 |
| RandomNeg | CRF | 0.477 | 0.427 | 0.451 |
| Retrieval | CRF | 0.536 | 0.356 | 0.428 |
| All | CRF | 0.527 | 0.346 | 0.418 |
| Positive | LSTM | 0.386 | 0.422 | 0.403 |
| RandomNeg | LSTM | 0.408 | 0.386 | 0.399 |
| Retrieval | LSTM | 0.464 | 0.328 | 0.384 |
| All | LSTM | 0.492 | 0.371 | 0.423 |

Table 2: Isnad retrieval results tested on **all documents** (2a, left) and on **retrieved documents** (2b, right)

## 5 Experiments

### 5.1 Training Setup Evaluation

First, we present our results evaluating the performance of different training and test setups on the obituary and isnad point detection tasks. All results we present in this section are trained with tenfold cross-validation. Recall that the documents used in the obituary experiments are chunked into 100-token pieces for the purposes of retrieval and tagging, of which 1.2% contain start points of obituaries, giving this dataset a very low base rate. The isnad dataset, in contrast, is broken into 75-token chunks and has a higher base rate of 21.1%. Also note that the model trained on retrieved documents is only used to label test documents that are also retrieved by the retrieval model used to select training examples, while in all other training setups the trained models is used to label every document in the test collection.

**Error! Reference source not found.**a above presents our results for the obituary extraction models, reporting the precision, recall, and F1 for obit starting points averaged across documents. As a baseline, training on positive examples only causes the model to hallucinate many false positives in documents where no obituaries exist, leading to poor performance for both the LSTM and CRF with local features. This issue is slightly alleviated with random negative sampling, as the model has been shown some examples of what not to label when trying to find where obituaries begin. However, the small increase in performance indicates that the examples are not very informative. Training on retrieved documents looks like it improves performance greatly, to the point where the CRF trained on retrieved data outperforms that trained on all the data. The slight performance gain for the LSTM may be due to the obvious negative examples still providing an opportunity for the model to tune the word embeddings. The high performance of the retrieval-based models, however, may be due to the fact that the retrieval model is only used to tag retrieved test documents rather than the full test set.

When the base rate of the phenomenon of interest is higher, as with the *isnad* tagging results in Table 2a, we see a similar pattern in the results, albeit with lower variance across training setups. The overall lower performance is indicative of the more difficult nature of detecting isnads as opposed to detecting obituaries.

As with the obituaries, the model trained on positive examples only performs the worst, with the addition of random negative samples slightly improving performance. Interestingly, the retrieval model performs only marginally better than random negative samples, at the cost of slightly decreased recall in exchange for a substantial increase in precision. As with the obituaries, it's unclear how much of this performance gain is a consequence of applying the model to retrieved documents as opposed to doing so on all documents, which we will explore in more detail below. Importantly, the variance in performance across training setups is much lower than for the obituaries. One might conclude that, as the set of potential negative samples shrinks relative to the size of the entire dataset (i.e. the base rate increases), the performance gain from random samples approaches that of samples curated by a retrieval system, as the two sampling methods will have more similar results when there are fewer negative samples to choose from.

The choice of training data also has a significant effect on the time it takes to train these models. This is especially significant for the Richmond Dispatch data, as the phenomenon of interest is extremely rare and the dataset is larger, so the difference between training on all the data and training on a smaller dataset is amplified. Training on all the data took around a day, while training on the subsampled datasets took a few hours. The OpenITI dataset is significantly smaller, so the difference in training time is less pronounced.

### 5.2 Retrieval Model Impact

Since the models discussed in the previous section were mostly tested on the full set of retrieved documents, this leaves open the question of how much of the performance gain on the model trained on retrieved documents comes from the choice of training data and how much is due to the choice of documents to tag at test time. To understand this more fully, we took the same model as trained in Section 5.1 and evaluated them in the setting in which they were only used to tag only documents retrieved by the retrieval system used to select test data for the retrieval-based system's evaluation. The results for the obit dataset can be seen in Table 1b above. Here, we see that most of the performance gains come from the lack of false positive in non-retrieved documents, indicating the selecting the documents you will apply the model to at test time is just as, if not more, important than selecting the documents used to train the model. Different methods of selecting training data seem to cause very small, likely insignificant in some cases, changes in performance when the test documents are restricted to the smaller pool of retrieved candidates. Somewhat oddly, there seems to be a precision-recall tradeoff inherent in choosing to train on random negative samples as opposed to training on the output of a retrieval model, with a slight increase in precision and a corresponding decrease in recall. The same trends are evident in the *isnad* results given in Table 2b above.

| Training Set | Model | P | R | F |
|---|---|---|---|---|
| Positive | CRF | 0.469 | 0.515 | 0.491 |
| RandomNeg | CRF | 0.567 | 0.601 | 0.584 |
| Retrieval | CRF | 0.905 | 0.875 | 0.890 |
| All | CRF | 0.874 | 0.835 | 0.854 |
| Positive | LSTM | 0.450 | 0.493 | 0.470 |
| RandomNeg | LSTM | 0.559 | 0.594 | 0.576 |
| Retrieval | LSTM | 0.915 | 0.866 | 0.890 |
| All | LSTM | 0.921 | 0.986 | 0.952 |

Table 3: Obit tagging results testing on **all documents** for the 10% base rate corpus

| Training Set | Model | P | R | F |
|---|---|---|---|---|
| Positive | CRF | 0.892 | 0.902 | 0.897 |
| RandomNeg | CRF | 0.897 | 0.898 | 0.897 |
| Retrieval | CRF | 0.905 | 0.875 | 0.890 |
| All | CRF | 0.911 | 0.851 | 0.880 |
| Positive | LSTM | 0.902 | 0.897 | 0.899 |
| RandomNeg | LSTM | 0.906 | 0.904 | 0.905 |
| Retrieval | LSTM | 0.915 | 0.866 | 0.890 |
| All | LSTM | 0.933 | 0.890 | 0.911 |

Table 4: Obit tagging results testing on **retrieved documents** only for the 10% base rate corpus

As with the choice of training data, the choice of test data also has a significant impact on runtime, which can be critical in some applications working with very large datatsets. It is often less expensive, time-wise, to run a first-pass retrieval model, then perform a more expensive inference procedure on the retrieved results, rather than performing inference on all documents at test time, which should be considered alongside any potential performance gains from tagging only retrieved documents.

## 5.3 Effect of Base Rate

Having explored the effects of different training and test setups on detecting rare events, we will now turn our attention to examining the effect of the base rate of the phenomenon in question. This is somewhat difficult to study with the *isnad* dataset, as way in which the ground truth data was collected makes it difficult to artificially inflate the base rate by discarding documents as there are few complete documents to set aside containing no *isnads*. However, in the case of the Richmond corpus, downsampling the negative class is much simpler as the vast majority (97 percent) of newspaper columns contain no obituaries. Thus, by downsampling the columns without obituaries, we can create a dataset with a much higher chunk-level base rate. For these experiments, we downsample the Richmond corpus such that we have a base rate of 10 percent, ten times higher than the full collection, and train models under the same training and test conditions as described in sections 5.1 and 5.2. The results of those experiments can be seen in Table 3 above. Since the corpus has been heavily altered, we use a new retrieval model for this corpus's new optimal operating point, using TFIDF-weighted unigram features, 100-token chunks, and a retrieval threshold of .14. The resultant retrieval model has an average precision of .79, so that is the precision we use for the random negative sampling training. As with the experiments above, the retrieval model is still tested only on retrieved documents.

These results will be compared against those in Table 4, where we only tag retrieved documents rather than all documents. As one might expect, the performance gap between training on positive documents only and other forms of training in the setting where all test documents are tagged is lower, since there are significantly fewer documents to have false positives in when the corpus is 90 percent smaller. Additionally, despite the higher base rate, there is still a significant difference between tagging retrieval output versus tagging. This makes intuitive sense given that the *isnad* dataset, with a base rate of around .2, still displays a difference in performance across test setups, implying that the cutoff for where it becomes more effective to tag all documents as opposed to using a retrieval system to tag likely documents should be above that point.

We see that for low base-rate settings, choosing what to evaluate at test time may be as important as choosing what to train the model on. However, there seems to be a precision-recall tradeoff involved in the choice of training data. It is unclear exactly what is causing the models trained on different datasets to gain precision but lose recall. It may be that a different form of evaluation, like performing a ranking evaluation of the output of the extraction models, might be more informative of exactly what's going on with these models. It may be the models trained on positive documents only are more likely to allow false positives due to the threshold of the extractor being poorly tuned, while models that have seen more negative examples are less likely to predict the phenomenon of interest due to a lower prior probability for the class.

## 6 Conclusions

We find that different methods of selecting training data bring tradeoffs in the precision and recall of the resultant models. While adding more difficult training examples or simply training on all the data can lead to overall increased performance, this often comes in the form of an increase in precision at the expense of recall. Regarding the influence of the base-rate, we see that the gap in performance between different training and test setups decreases as the base rate increases, although the same precision-recall tradeoff still applies. Using the datasets we're working with, it is difficult to examine very high base-rate settings, so we are unable to find a clear crossover point where testing on all data is the clearly better choice. As such, it would be valuable to examine the influence of the base rate using a dataset where we have better control over the exact base rate of the phenomenon of interest. We've established a lower bound on the point where testing on all data makes more sense, but in order to explore that more fully, new data that can be more easily manipulated to give a desired base rate will be needed. It would also be worthwhile to investigate if the same tradeoffs in performance persist in other domains.